\documentclass[10pt,twocolumn]{article}
\usepackage{cite}
\usepackage{amsmath,amssymb,times,amsfonts}
\usepackage{algorithmic}
\usepackage{graphicx}
\usepackage[dvipsnames]{xcolor}
\usepackage{booktabs}
\usepackage[hidelinks, bookmarks=false]{hyperref}
\hypersetup{
    colorlinks=true, 
    citecolor=blue,  
    linkcolor=blue,  
    urlcolor=blue,   
    pdfborder={0 0 0} 
}

\usepackage{pifont}
\newcommand{\cmark}{\ding{51}}%
\newcommand{\xmark}{\textcolor{lightgray}{\ding{55}}}%

\newcommand{\bm}[1]{\mathbf{#1}} 

\newcommand\T{{\mathpalette\raiseT\intercal}}
\newcommand\raiseT[2]{%
\setbox0\hbox{$#1{#2}$}\raise\dp0\box0}

\date{}

\begin{document}

\title{\textbf{AdaKAN: A Dual-Branch Adaptive Kolmogorov-Arnold Network for Medical Image Segmentation}}
\author{}

\author{Dalia Alzu'bi, Deep Bhattacharyya, Ali Ayub, and A. Ben Hamza\\
Department of Cybersecurity and Intelligent Systems Engineering\\
Concordia University, Montreal, Canada}

\maketitle

\begin{abstract}
Medical image segmentation is a fundamental task in computer-aided diagnosis, yet it remains challenging due to the complexity of anatomical structures and the variability across imaging modalities. In this paper, we propose AdaKAN, an Adaptive Kolmogorov-Arnold Network (KAN) that synergistically integrates convolutional operations with a novel efficient KAN (EffiKAN) block, comprised of an efficient attention mechanism and an adaptive KAN (AdaptKAN) module. This module features a dual-branch design: one branch employs a KAN layer with Bernstein polynomial activations for globally smooth and stable function approximation, while the other branch performs channel-wise refinement through projection operations and adaptive scaling. AdaKAN adopts a U-shaped architecture that effectively captures both long-range dependencies and fine-grained local features, overcoming the limitations of conventional convolutional and Transformer-based segmentation models. Skip connections are employed to preserve spatial details during encoding and facilitate accurate reconstruction during decoding. Extensive experiments conducted on diverse medical imaging datasets demonstrate that AdaKAN achieves state-of-the-art performance in segmentation accuracy. Code is available at: \textcolor{blue}{https://github.com/bhattacharyyadeep/KAT-Project}
\end{abstract}


\section{Introduction}
Medical image segmentation aims to identify and delineate anatomical structures and pathological regions from diverse imaging modalities. Fully Convolutional Networks (FCNs) play a central role in many contemporary medical image segmentation approaches. U-Net based architectures~\cite{Khader2025AGU-Net} leverage hierarchical convolutional feature extraction within a symmetric encoder-decoder framework to model local spatial patterns effectively. While FCN-based architectures have demonstrated significant success, they are fundamentally constrained by the intrinsic locality of convolutional operations. This restricted receptive field limits their ability to model the long-range spatial dependencies, often resulting in blurred boundaries or incomplete segmentation of complex structures. To overcome these limitations and better capture global contextual information, Transformer-based architectures ~\cite{Val2021MedT,Noh2025Dual,ChenTransUNet} have been proposed, leveraging self-attention mechanisms to model long-range dependencies. While effective, their computational complexity is quadratic. On the other hand, lightweight MLP-based architectures~\cite{ValanarasuUNeXt,RuanMALUNet} emphasize computational efficiency, but often suffer from limited expressive capacity, which can compromise the modeling of complex anatomical structures.

Kolmogorov-Arnold Networks (KANs)~\cite{Liu2024KAN} have recently gained attention as an alternative to MLPs, shifting the learnable parameters from fixed-activation nodes to edge-based univariate activation functions, improving interpretability and enabling modeling of complex, non-linear relationships. Building upon this foundation, U-KAN~\cite{li2024U-KAN}  integrates B-spline-based KAN layers within a U-shaped architecture, but exhibits limited capacity for modeling global context, which is critical for accurate segmentation of anatomical structures. The above-mentioned limitations underscore the need for a balanced framework that unifies local precision, global awareness, and computational efficiency across diverse medical imaging scenarios. To this end, we introduce AdaKAN, an adaptive segmentation framework that integrates efficient attention with the adaptive KAN module to enable expressive yet computationally efficient modeling. The adaptive KAN module adopts a dual-branch architecture: the left branch incorporates a KAN layer with Bernstein polynomial activations, offering globally smooth and stable function approximation; the right branch performs channel-wise refinement via  a bottleneck structure comprised of down- and up-projections, and modulated by a learnable coefficient. Our main contributions can be summarized as follows:

\begin{itemize}
\item We propose AdaKAN, an encoder-decoder network mixing convolutional and KAN stages.
 \item We introduce an AdaptKAN module, which uses a Bernstein polynomial-based KAN layer.
 \item Experimental results show that AdaKAN achieves superior segmentation performance.
\end{itemize}

\section{Related Work}
Fully Convolutional Networks (FCNs) have long served as the cornerstone of medical image segmentation due to their ability to preserve spatial hierarchies via pixel-wise prediction. Among attention-augmented FCNs, Attention U-Net~\cite{OktayAttentionU-Net} introduces gated signal propagation to suppress irrelevant regions and emphasize salient features during upsampling. Despite their strength in local feature extraction and hierarchical modeling, FCN-based approaches remain limited in their ability to model long‑range spatial dependencies, as convolutional operations primarily capture local contextual information. Inspired by the success of Vision Transformers (ViT)~\cite{DosovitskiyViT}, recent literature has witnessed a paradigm shift toward Transformer-based architectures, which primarily aim to mitigate the inherent inductive biases of FCNs. TransUNet~\cite{ChenTransUNet} fuses Transformer blocks within a standard U-Net bottleneck, employing self-attention and cross-attention to maintain high-resolution spatial details while encoding global context. MLP-based methods offer further efficiency gains~\cite{ValanarasuUNeXt,LiuRolling-Unet}. Despite their computational advantages, both Transformer- and MLP-based methods often trade off expressive capacity or require extensive pre-training.

More recently, KANs~\cite{Liu2024KAN} have emerged as a powerful alternative to MLPs, leveraging compositions of learnable univariate functions with enhanced interpretability and flexibility. Building upon this foundation, U-KAN~\cite{li2024U-KAN} adapts the U-Net architecture by replacing convolutional layers with tokenized KAN blocks using B-spline activations, achieving improved segmentation accuracy. However, existing KAN-based methods often rely on B-spline parameterizations, which, while smooth, limit adaptive expressiveness and struggle to efficiently integrate long-range contextual modeling with lightweight computation. In contrast, our AdaKAN framework introduces the EffiKAN block, integrating Efficient Attention with the Adaptive Kolmogorov-Arnold Network (AdaptKAN). Within this block, the KAN layer employs Bernstein polynomials as its basis activation, enabling enhanced global adaptability and smooth functional approximation. This design allows the network to effectively balance global context modeling with local feature refinement, fostering robust information propagation across encoder-decoder stages. By integrating these components within a symmetric encoder-decoder architecture, AdaKAN effectively balances segmentation accuracy and computational efficiency.

\section{Method}
In this section, we formulate the semantic segmentation task, and provide a preliminary background on KANs. Then, we present an overview of the proposed model architecture, and describe in detail its main architectural components.

\subsection{Preliminaries}
\noindent\textbf{Problem Statement.}\: Given an input image $\bm{I}\in \mathbb{R}^{H \times W \times 3}$ with height $H$ and $W$ width, our aim is to generate a pixel-wise segmentation mask $\bm{M}\in [C]^{H \times W}$, where $[C]$ denotes the discrete set $\{0,\dots,C-1\}$ of $C$ semantic classes. Each pixel in $\bm{M}$ is assigned an integer label (i.e., semantic class) from 0 to $C-1$.

\medskip\noindent\textbf{Kolmogorov-Arnold Networks.}\: The KAN layer is an integral element of KANs~\cite{Liu2024KAN}, represented by a matrix $\mathbf{\Phi}=(\phi_{q,p})$ of one-dimensional functions, where each trainable activation function $\phi$ is expressed as $\phi(x) = w_{b}\text{SiLU}(x) + w_{s}\text{spline}(x),$ where $\text{spline}(x) = \sum_{i}c_{i}B_{i}(x)$ represents a weighted sum of B-spline basis functions with learnable coefficients $c_{i}$. During the training process, the weights $w_{b}$ and $w_{s}$ are optimized to enhance performance. For an input $\mathbf{x}^{(l)}\in\mathbb{R}^{F_{l}}$, the output of the KAN layer is $\mathbf{x}^{(l+1)} = \text{KAN}(\mathbf{x}^{(l)})$.

\begin{figure*}[!ht]
\begin{center}
\includegraphics[scale=0.65]{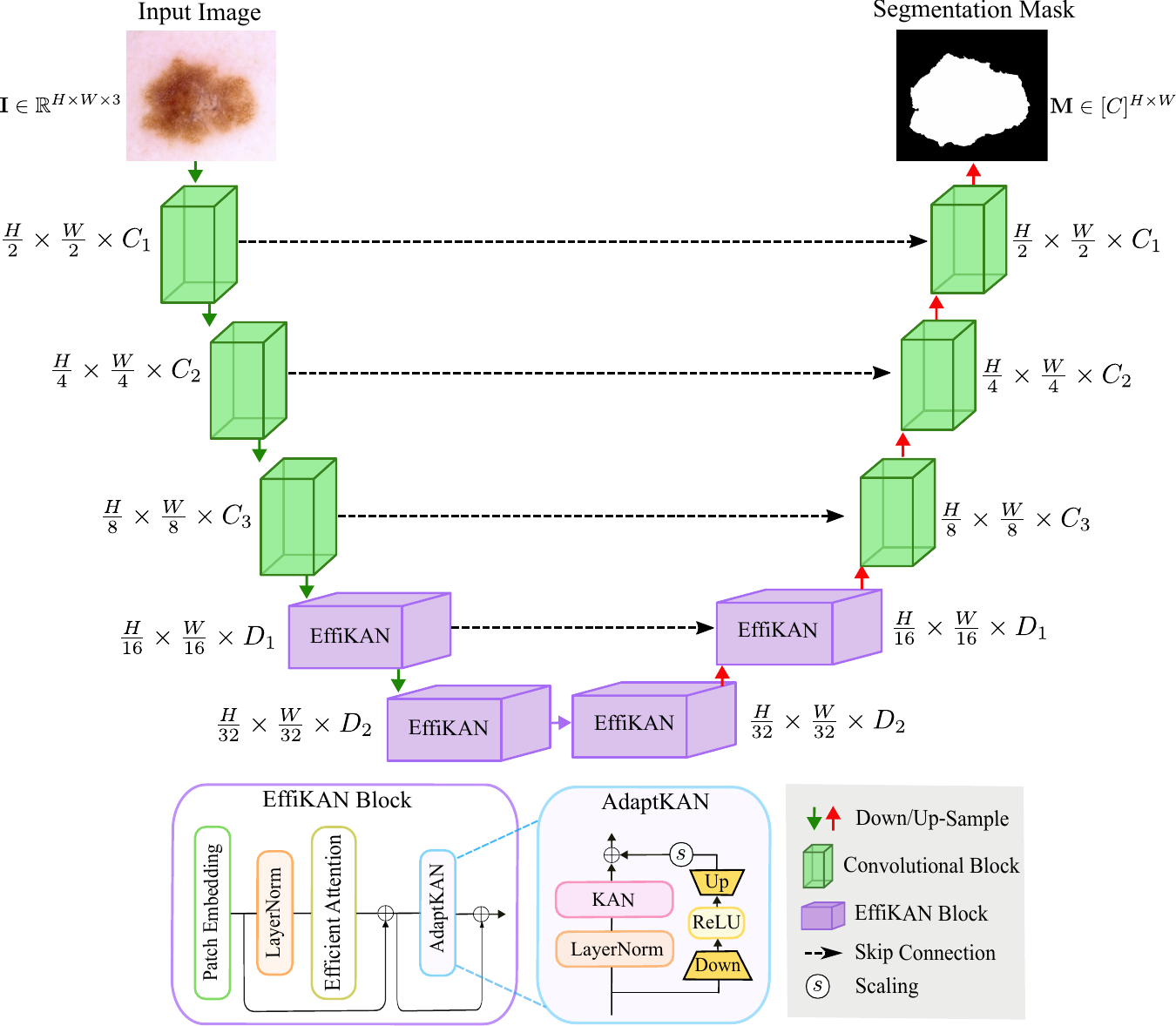}
\end{center}
\caption{\textbf{Overview of AdaKAN framework.} The proposed model adopts a symmetric encoder-decoder architecture, comprising a convolution stage and an efficient KAN stage. The encoder downsamples the input image through three convolution blocks followed by two EffiKAN blocks, progressively reducing spatial resolution while expanding feature channels to capture hierarchical representations. The decoder mirrors this design, upsampling features via bilinear interpolation and integrating high-resolution details from the encoder via skip connections to produce the segmentation mask.}
\label{Fig:ModelArchitecture}
\end{figure*}

\subsection{Model Architecture}
Figure~\ref{Fig:ModelArchitecture} illustrates the overall structure of the AdaKAN framework, which features a symmetric U-shaped encoder-decoder structure incorporating convolutional operations, an efficient attention mechanism and an adaptive function approximation to effectively capture both local and global contextual features. The process comprises two main stages: a convolutional stage and an EffiKAN stage. The convolution stage, with three consecutive blocks, is designed to progressively reduce spatial resolution while learning multi-scale low-level semantic features. Given the input image $\bm{I}\in \mathbb{R}^{H \times W \times 3}$, three consecutive convolutional blocks progressively down-sample the feature maps to $H/2\times W/2\times C_{1}$, then to $H/4\times W/4\times C_{2}$, and finally to $H/8\times W/8\times C_{3}$.  These features are then passed to the EffiKAN stage, which introduces the novel EffiKAN block, a fusion of efficient attention and the adaptive KAN (AdaptKAN) module. Efficient attention enables global feature interaction with linear computational complexity, overcoming the limitations of standard dot-product self-attention mechanisms. The AdaptKAN module, central to AdaKAN’s adaptability, comprises two parallel branches: the left branch applies a KAN layer with Bernstein polynomial-based activation functions to achieve globally smooth and stable approximations, while the right branch performs channel-wise refinement through down-projection, ReLU activation, and up-projection, modulated by a learnable scaling coefficient. The outputs of these branches are then added, allowing the model to balance global context modeling with local precision. The decoder mirrors the encoder structure, using bilinear upsampling and skip connections to restore spatial resolution and refine features. The final segmentation mask is produced through a $1\times 1$ convolution followed by softmax activation, enabling pixel-wise classification. Overall, AdaKAN's architecture is designed to be both expressive and efficient, making it well-suited for medical image segmentation tasks.

\subsection{Encoder}
The encoder is composed of two main stages:

\smallskip\noindent\textbf{Convolutional Stage.}\: This stage is composed of $L$ convolutional blocks, where each unit sequentially integrates a $3\times 3$ convolution (Conv) layer, batch normalization (BN), and a rectified linear unit (ReLU) activation function. The process of this block can be expressed as:
\begin{equation}
\bm{X}^{(\ell)} = \text{ReLU}(\text{BN}(\text{Conv}(\bm{X}^{(\ell-1)}))),
\label{Eq:ConvSE}
\end{equation} 
with $\bm{X}^{(0)} = \bm{I}$ being the input image. After each block, a $2\times 2$ max-pooling layer is applied to downsample the feature map. The output feature map $\bm{X}^{(L)}$ from the last convolution block is fed into the first block of the subsequent efficient KAN stage.

\medskip\noindent\textbf{Efficient KAN Stage.}\: This stage consists of $K$ EffiKAN blocks, each featuring a patch embedding layer, a layer normalization, an efficient attention mechanism, and an adaptive KAN (AdaptKAN) module. Residual connections are also incorporated to retain and reuse low-level features from earlier layers.

\smallskip\noindent\textit{Patch Embedding.}\: The output from the $L$-th convolutional block is the feature map $\bm{X}^{(L)} \in \mathbb{R}^{\frac{H}{2^L}\times\frac{W}{2^L}\times C_L}$, which serves as input to the first EffiKAN block. It is converted into a tokenized feature matrix $\bm{Y} \in \mathbb{R}^{n\times d}$ using a patch embedding layer, where $n$ is the total number of patches and $d$ is the embedding dimension. Adopting the structural principles of the Vision Transformer~\cite{DosovitskiyViT}, we partition the input feature map into non-overlapping flattened patches and then project it into an embedding space via a learnable linear projection.

\smallskip\noindent\textit{Efficient Attention.}\: The tokenized feature matrix $\bm{Y}$ first passes through Layer Normalization (LN), which ensures stable distribution of features keeping the same dimension, yielding the normalized feature matrix $\text{LN}(\bm{Y})\in\mathbb{R}^{n\times d}$ that serves as input to the efficient attention layer~\cite{Shen2021EfficientAttention}. This layer captures global feature dependencies with linear computational complexity, in contrast to the quadratic computational complexity of standard attention mechanisms. Specifically, the normalized feature matrix $\text{LN}(\bm{Y})$ is first linearly projected into the query matrix $\bm{Q} \in \mathbb{R}^{n \times d_k}$, key matrix $\bm{K} \in \mathbb{R}^{n \times d_k}$, and value matrix $\bm{V} \in \mathbb{R}^{n \times d}$, where $d_k$ is the projection dimension. Then, the output of the efficient attention (EA) layer is computed as follows:
\begin{equation}
\text{EA}(\text{LN}(\bm{Y})) = \sigma_{\text{row}}(\bm{Q})(\sigma_{\text{col}}(\bm{K})^{\T}\bm{V}),
\end{equation}
where $\sigma_{\text{row}}$ and $\sigma_{\text{col}}$ denote row- and column-wise softmax functions, respectively. Subsequently, the output of the efficient attention layer is fused with the tokenized feature matrix via a residual connection to produce a new feature matrix:
\begin{equation}
\bm{Z} =  \text{EA}(\text{LN}(\bm{Y})) + \bm{Y} ,
\end{equation}
where $\bm{Z} \in \mathbb{R}^{n \times d}$ serves as input to the adaptive KAN (AdaptKAN) module. In the efficient attention layer, each output token depends on a weighted combination of all value tokens, because $\sigma_{\text{col}}(\bm{K})^{\T}\bm{V}$ aggregates information across the entire token set (all spatial patches). This preserves the key property of attention: each position can incorporate information from any other position in a single layer, enabling global context modeling. On the other hand, the standard dot-product attention forms the matrix $\bm{Q}\bm{K}^{\T}$ of size $n\times n$, which costs $\mathcal{O}(n^2)$ time/memory. Efficient Attention reorders multiplications: it first computes $\sigma_{\text{col}}(\bm{K})^{\T}\bm{V}$, which is $\mathcal{O}(n)$ in tokens (given fixed channel dimensions), and then multiplies by $\sigma_{\text{row}}(\bm{Q})$, again $\mathcal{O}(n)$. Thus, the dominant cost becomes linear in $n$.

\medskip\noindent\textit{AdaptKAN Module.}\: This is a key component of the AdaKAN architecture, designed to enhance representational power while preserving computational efficiency. It consists of two parallel branches. The left branch integrates layer normalization and a KAN Layer with Bernstein polynomials as its basis activation functions, while the right branch is a learnable bottleneck structure comprised of a down-projection layer to reduce the dimension, followed by ReLU activation, then an up-projection layer to restore the dimension, together with a learnable adaptive scaling coefficient applied to the bottleneck's output.  This learned scaling factor modulates the contribution of the right branch, providing a control mechanism over the influence of the learnable lightweight bottleneck structure on the output, relative to the much larger KAN layer. The output of the AdaptKAN module is computed as the sum of the KAN layer output and the scaled bottleneck structure output. Specifically, the Bernstein polynomial-based activation function is defined as
\begin{equation}
\text{Bern}(z) = \sum_{r=0}^{R} \theta_{r} P_{r,R}(z) = \sum_{r=0}^{R} \theta_{r} {R\choose r}z^{r}(1-z)^{R-r},
\end{equation}
where $P_{r,R}(z)$ is the $r$-th Bernstein basis polynomial of order $R$ and $\theta_r$ are learnable coefficients. The KAN layer is defined over an interval $[0,1]$ using a weighted sum of a SiLU activation function and a Bernstein polynomial expansion:
\begin{equation}
\tilde{\phi}(z)=\tilde{w}_{b}\,\text{SiLU}(z) + \tilde{w}_{s}\,\text{Bern}(z).
\label{Eq:Bernstein}
\end{equation}
Bernstein polynomials, globally defined, can approximate any continuous function on an interval of $[0,1]$. The process of the AdaptKAN module can be expressed as:
\begin{equation}
\text{AdaptKAN}(\bm{Z}) = \text{KAN}(\text{LN}(\bm{Z}))+ s \cdot {\tilde{\bm{Z}}},
\end{equation}
where
\begin{equation}
\bm{\tilde{Z}} = \text{ReLU}\big(\bm{Z} \bm{W}_{\text{down}}\big) \bm{W}_{\text{up}}
\end{equation}
is the intermediate feature matrix obtained from the right branch. The parameters $\bm{W}_{\text{down}}$ and $\bm{W}_{\text{up}}$ correspond to learnable down-projection and up-projection weight matrices, respectively, which are applied sequentially with a ReLU activation to capture nonlinear channel dependencies. The learnable adaptive scaling coefficient $s$ modulates the contribution of the right branch before the outputs of both branches are added. Without this scaling factor, the lightweight bottleneck structure could initially produce large or poorly scaled activations. Adding this to the output of the KAN layer could disrupt the learned representation and destabilize the training process. Hence, the scaling factor $s$ is an essential control knob that manages the trade-off between preserving the expressive KAN knowledge and incorporating task-specific learned features. 

\smallskip\noindent The proposed AdaptKAN module improves representational power and preserves efficiency. The Bernstein-based KAN branch provides globally smooth approximation on $[0,1]$, while the bottleneck branch performs lightweight channel-wise refinement. For representational power, using a Bernstein expansion inside the KAN edge activations provides a globally smooth and stable function approximation mechanism. This supports modeling of smooth anatomical shape variations across images. Also, KANs move learnable capacity into edge functions (univariate learnable activations) rather than fixed node activations, which improves flexibility/interpretability. For efficiency, the bottleneck refinement is lightweight, as the right branch is a low-rank channel mixer: down-project to a smaller dimension and up-project back, so its parameter and compute cost scale with the bottleneck width rather than full width. This bottleneck design is a standard way to introduce extra nonlinearity without expensive full MLP blocks. Moreover, the bottleneck branch is deliberately gated by a learnable scaling coefficient $s$ that stabilizes fusion. This learnable scalar is initialized to a small value (e.g., 0.1) so that training starts close to the KAN branch output and gradually learns how much the bottleneck refinement should contribute.

\smallskip\noindent By integrating efficient attention and adaptive KAN mechanisms, the proposed EffiKAN block is able to capture both local and global dependencies in a computationally efficient manner. Patch embedding and layer normalization provide a stable input representation, efficient attention ensures long-range contextual reasoning at linear-time complexity, and AdaptKAN enriches the representation with global function approximation and multi-scale feature fusion. This makes the EffiKAN block well-suited for high-resolution segmentation tasks. Hence, given a tokenized input feature matrix $\bm{Y}$, the output of the EffiKAN block can be expressed as:
\begin{equation}
\text{EffiKAN}(\bm{Y}) = \text{AdaptKAN}(\bm{Z}) + \bm{Z}.
\end{equation}
The bottleneck acts as the bridge between the encoder and decoder, enabling deep feature learning before the upsampling process in the decoder begins.

\subsection{Decoder}
The decoder also consists of two main stages, an EffiKAN stage and a convolutional stage, following a structure that mirrors the encoder for architectural coherence. In the EffiKAN stage, each block begins by upsampling the input feature maps via bilinear interpolation by a factor of two, facilitating a progressive restoration of spatial resolution. These upsampled representations are subsequently integrated with the corresponding encoder features through skip connections, which are instrumental in recovering high-frequency spatial details often attenuated during the downsampling process. This architectural choice effectively bridges the semantic gap between the contracting and expanding paths, thereby enhancing boundary delineation and localization precision. The final reconstruction phase consists of three additional upsampling blocks that iteratively refine the feature maps into the final output. Each block employs bilinear interpolation and concatenation with encoder features, progressively enhancing the spatial and semantic consistency of the decoded maps. The segmentation mask is obtained by applying a $1\times1$ convolution with $C$ channels to project the refined features into the class space, followed by a softmax function to produce pixel-wise class probabilities. 

\section{Experiments}
\begin{table*}[!htb]
\caption{Performance comparison of our method and baselines on the BUSI, GlaS, and CVC datasets. The top-performing and second-best results are highlighted in bold and underlined, respectively.}
\smallskip
\setlength\tabcolsep{5.8pt}
\centering
\footnotesize
\begin{tabular}{@{}lcccccccccc}
\toprule
& & & \multicolumn{2}{c}{BUSI} & \multicolumn{2}{c}{GlaS} & \multicolumn{2}{c}{CVC} \\
\cmidrule(lr){4-5} \cmidrule(lr){6-7} \cmidrule(lr){8-9}
Method & Params(M)$\downarrow$ & FLOPS(G)$\downarrow$ & IoU$\uparrow$ & DSC$\uparrow$ & IoU$\uparrow$ & DSC$\uparrow$ & IoU$\uparrow$ & DSC$\uparrow$ \\
\midrule
Att-UNet~\cite{OktayAttentionU-Net} &50.27 & 60.06 & 55.18$\pm$3.61 & 70.22$\pm$2.88 & 86.84$\pm$1.19 & 92.89$\pm$0.65 & 84.52$\pm$0.51 & 91.46$\pm$0.25 \\
MedT~\cite{Val2021MedT} &\underline{1.60} & 21.24 & 52.15$\pm$3.47 & 67.68$\pm$3.18 & 75.47$\pm$3.46 & 85.92$\pm$2.93 & 60.08$\pm$0.92 & 79.18$\pm$1.05 \\
UNeXt~\cite{ValanarasuUNeXt} & \textbf{1.47} & \textbf{4.58} & 61.78$\pm$1.46 & 75.52$\pm$0.91 & 83.95$\pm$1.09 & 91.22$\pm$0.67 & 74.83$\pm$0.24 & 85.36$\pm$0.17 \\
Rolling-Unet~\cite{LiuRolling-Unet} & 25.06 & 28.32 & 61.00$\pm$0.64 & 74.67$\pm$1.24 & 86.42$\pm$0.96 & 92.63$\pm$0.62 & 82.87$\pm$1.47 & 90.48$\pm$0.83 \\
U-KAN~\cite{li2024U-KAN} & 9.38 & \underline{6.89} & 63.38$\pm$2.83 & 76.42$\pm$2.90 & 87.64$\pm$0.32 & 93.37$\pm$0.16 & 85.05$\pm$0.53 & 91.88$\pm$0.29 \\
PDS-UKAN~\cite{Deng2025PDS} & -- & -- & 62.74$\pm$ --  & 76.99$\pm$ -- & \underline{88.55$\pm$ -- }& \underline{93.93$\pm$ -- } & -- & -- \\
ResU-KAN~\cite{Wang2025ResU-KAN} & 20.06 & 8.57 & \underline{67.74$\pm$1.35} & \underline{79.92$\pm$0.73} & 87.99$\pm$0.44& 93.61$\pm$0.24& \underline{85.38$\pm$0.87}& \underline{92.12$\pm$0.56} \\
\midrule
Ours & 14.21 & 8.12 & \textbf{68.84$\pm$1.04} & \textbf{80.95$\pm$0.75} & \textbf{88.58$\pm$0.21} & \textbf{93.94$\pm$0.13} & \textbf{85.73$\pm$0.49} & \textbf{92.26$\pm$0.29} \\
\bottomrule
\end{tabular}
\label{TAB:tab1}
\end{table*}

\subsection{Experimental Setting}
\noindent\textbf{Datasets.}\: We evaluate our proposed image segmentation model on five datasets: Breast Ultrasound Images (BUSI)~\cite{AlDhabyani2020}, GlaS~\cite{gland}, CVC~\cite{cvc}, ISIC 2018~\cite{codella2018skin}, and ACDC~\cite{bernard2018deep}.

\medskip\noindent\textbf{Evaluation Metrics.}\: We employ two established metrics: Intersection over Union (IoU) and the Dice Similarity Coefficient (DSC)  between the predicted segmentation mask ($P$) and the ground truth mask ($G$):
\begin{equation}
	\text{IoU} = \frac{|P\cap G|}{|P\cup G|}\quad\text{and}\quad \text{DSC} = \frac{2\times |P\cap G|}{|P| + |G|}.
\end{equation}

\medskip\noindent\textbf{Baselines.}\: We compare AdaKAN against several SOTA methods, including Att-UNet~\cite{OktayAttentionU-Net}, MedT~\cite{Val2021MedT}, UNeXt~\cite{ValanarasuUNeXt}, U-KAN~\cite{li2024U-KAN}, PDS-UKAN~\cite{Deng2025PDS}, and ResU-KAN~\cite{Wang2025ResU-KAN}.

\medskip\noindent\textbf{Implementation Details.}\: We perform our experiments on an NVIDIA A4500 GPU using PyTorch framework. A batch size of 8 is adopted for the BUSI, CVC, and ISIC 2018 datasets. The GlaS dataset is trained with a batch size of 6, and the multi-class ACDC dataset utilizes a batch size of 16. The input image is resized to $256 \times 256$ for the BUSI, CVC-ClinicDB, ISIC 2018, and ACDC (per slice) datasets, while the GlaS dataset images are resized to $512 \times 512$. We adopt a learning rate of $10^{-4}$. The channel configuration follows $C_1=32$, $C_2=64$, $C_3=256$, $D_1=320$, and $D_2=512$, with a Bernstein order $R=3$. Model training uses the Adam optimizer and cosine annealing, decaying the learning rate to a minimum of $10^{-5}$. We split each dataset into $80\%$ training and $20\%$ validation. For binary segmentation, the training objective is defined as the average of the BCE and Dice losses. For the ACDC dataset, we instead use the average of Cross‑Entropy and Dice losses. All models are trained for 400 epochs, and results are reported as the mean and standard deviation over five independent random seeds.

\subsection{Experimental Results and Analysis}
\noindent\textbf{Quantitative Results on BUSI, GlaS, and CVC.}\: Table~\ref{TAB:tab1} provides a comprehensive benchmarking of the proposed AdaKAN model against state-of-the-art (SOTA) methods across the BUSI, GlaS, and CVC benchmarks. The evaluation encompasses architectural efficiency, measured by model parameter count and computational complexity in FLOPs, alongside a quantitative assessment of segmentation performance. AdaKAN achieves relative improvements of 1.1\% and 1.03\% in IoU and DSC, respectively, on the BUSI dataset compared to the second-best method, ResU-KAN. It exhibits an improvement of 0.03\% and 0.01\% in IoU and DSC, respectively, on the GlaS dataset compared to PDS-UKAN, and 0.35\% and 0.14\% higher IoU and DSC, respectively, on the CVC dataset compared to ResU-KAN. AdaKAN also features a balanced parameter count and FLOPs.

\begin{table}[!htb]
\caption{Performance comparison of our method and baselines on the ISIC 2018 dataset.}
\smallskip
\label{TAB:isic}
\setlength{\tabcolsep}{34pt}
\centering
\footnotesize
\begin{tabular}{@{}lcc@{}}
\toprule
Method & IoU & DSC\\
\midrule
Att-UNet~\cite{OktayAttentionU-Net} & 78.43 & 87.91 \\
MedT~\cite{Val2021MedT} & 81.48 & 89.49 \\
UNeXt~\cite{ValanarasuUNeXt} & 81.70 & 89.70 \\
Rolling-Unet~\cite{LiuRolling-Unet} & 83.74 & 90.90 \\
SET~\cite{Wang2025SET} & \underline{84.64} & 90.98 \\
U-Net V2~\cite{PengU-NetV2} & 84.15 & \underline{91.52} \\
\midrule
Ours & \textbf{84.88} & \textbf{91.71} \\
\bottomrule
\end{tabular}
\end{table}

\medskip\noindent\textbf{Quantitative Results on ISIC.}\: Table~\ref{TAB:isic} reports the performance comparison of AdaKAN with baselines in terms of IoU and DSC on the ISIC 2018 dataset. AdaKAN achieves relative improvements of 0.24\% in IoU compared to the second-best method, SET, and 0.19\% in DSC compared to U-Net V2.

\begin{table}[!htb]
	\caption{Performance comparison of our model and baselines on the ACDC dataset. The last column reports the average DSC scores of RV, Myo, and LV.}
	\smallskip
	\setlength{\tabcolsep}{6pt}
	\label{TAB:acdc}
	\centering
\footnotesize
	\begin{tabular}{@{}lccccc@{}}
		\toprule
		& \multicolumn{3}{c}{ACDC} &  \\
		
		\cmidrule{2-4}
		Method & RV & Myo & LV & Avg (\%)\\
		\midrule
		QMaxViT-UNet+~\cite{Nguyen2025QMaxViT} & 84.50 & 86.80 & 91.90 & 87.60 \\
		ScribFormer~\cite{Li2024Scribformer}  & 84.50 & 84.80 & 90.30 & 86.50 \\
		SelfRegularized-UNet~\cite{Zhu2024SelfReg-UNet} & 88.92 & 89.49 & 95.88 & 91.43\\
		PVT-EMCAD-B2~\cite{Rahman2024EMCAD} & 90.65 & \underline{89.68} & \underline{96.02} & \underline{92.12} \\
		\midrule
		Ours & \textbf{93.47} & 89.09 & \textbf{96.43} & \textbf{93.00} \\
		\bottomrule
	\end{tabular}
\end{table}

\begin{figure*}[!htb]
	\centering
	\includegraphics[width=.85\linewidth]{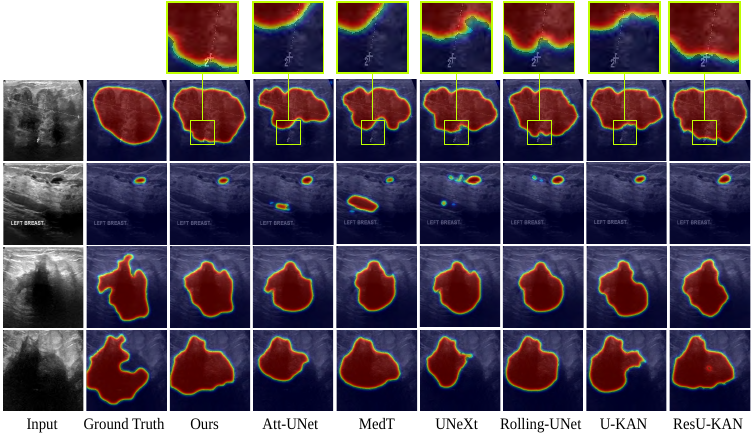}
	\caption{Visual comparison of segmentation heatmaps between the proposed AdaKAN and baselines on the BUSI dataset.}
	\label{FIG:BUSI}
\end{figure*}

\medskip\noindent\textbf{Quantitative Results on ACDC.}\: Table~\ref{TAB:acdc} presents a comparative analysis of AdaKAN against strong baselines on the ACDC dataset. Performance is quantified using the DSC metric across three anatomical cardiac structures: the Right Ventricle (RV), Myocardium (Myo), and Left Ventricle (LV). AdaKAN achieves a 0.88\% higher average DSC than PVT-EMCAD-B2.

\medskip\noindent\textbf{Qualitative Results.}\: Figure~\ref{FIG:BUSI} presents a qualitative comparison on four BUSI samples, showing the original input, the corresponding ground-truth heatmap, and the predicted heatmaps generated by AdaKAN alongside SOTA baselines. For the first sample, most models introduce notable segmentation errors. While ResU-KAN performs competitively, AdaKAN is more accurate, avoiding the over-segmentation observed in ResU-KAN. The second sample reveals that several SOTA baselines produce false positive foreground regions, whereas KAN-based models avoid such artifacts. Even then, U-KAN shows noticeable under-segmentation, while ResU-KAN and AdaKAN yield more comparable outputs. In the third and fourth samples, both characterized by pronounced irregularity, AdaKAN consistently delivers sharper boundary localization and lower segmentation error than competing approaches, including other KAN variants. This highlights the robustness of AdaKAN in handling noisy ultrasound images and preserving complex anatomical shapes.

\begin{figure}[!htb]
	\centering
	\includegraphics[width=.99\linewidth]{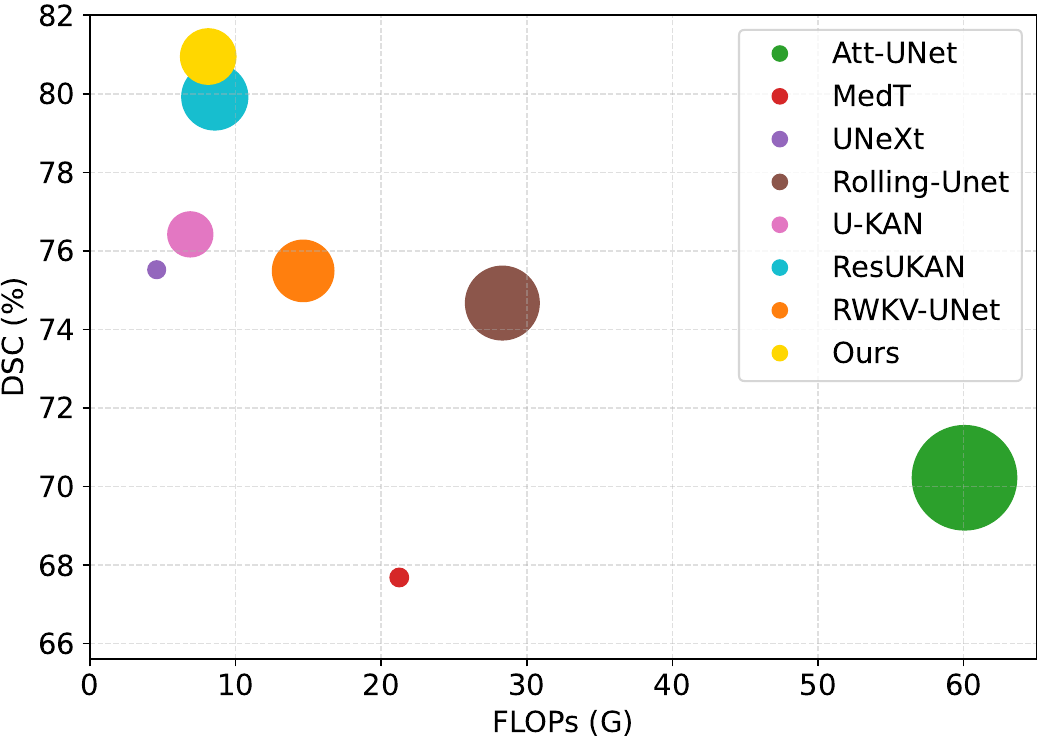}
	\caption{Comparison of model efficiency and performance of AdaKAN with baselines in terms of DSC, FLOPs, and parameter count (in millions) on BUSI. The lower number of FLOPs and parameter count indicates better efficiency.}
	\label{FIG:Flops}
\end{figure}

\smallskip\noindent\textbf{Model Efficiency Analysis.}\: Figure~\ref{FIG:Flops} illustrates the relationship between DSC, FLOPs, and model parameter count for AdaKAN in comparison with SOTA baselines on the BUSI dataset. The size of each circular marker denotes the model’s parameter count, providing an intuitive overview of computational cost and model complexity. The yellow circle corresponds to AdaKAN, which attains the highest DSC of 80.95\% while maintaining only 14.21M parameters and 8.12G FLOPs. Although ResU-KAN achieves a comparable DSC, its substantially larger parameter count (20.06M) makes it less computationally efficient. UNeXt, represented by a smaller purple circle, comprises fewer parameters, but
exhibits a much lower DSC compared to AdaKAN. Overall, the comparison highlights AdaKAN's superior efficiency-accuracy balance compared to SOTA baselines.

\medskip\noindent\textbf{Ablation Study.}\: Table~\ref{Table:EffikanComponents} evaluates the individual contributions of the core architectural components to the overall performance using the BUSI dataset. The empirical results demonstrate that the full configuration achieves optimal segmentation accuracy. Notably, the removal of the Efficient Attention layer results in a significant performance degradation,  underscoring the layer's critical function in capturing the global dependencies essential for accurate structure delineation. Similarly, removing the AdaptKAN module leads to IoU and DSC scores dropping by 4.75\% and 3.44\%, respectively. 

\begin{table}[!htb]
	\caption{Impact of key components on AdaKAN's performance evaluated on the BUSI dataset. }
	\centering
	\small
	\setlength{\tabcolsep}{1.4pt}
	\footnotesize
	\begin{tabular}{@{}c c c c c}
		\toprule
		\multicolumn{3}{c}{Model Components} & \multicolumn{2}{c}{Performance (\%)} \\
		\cmidrule(lr){1-3} \cmidrule(lr){4-5}
		Efficient Attention & Up-Projection & Down-Projection & IoU  & DSC  \\
		\midrule
		\cmark & \cmark & \cmark & \textbf{68.84} & \textbf{80.95} \\
		\cmark & \cmark & \xmark & 65.28 & 78.54 \\
		\cmark & \xmark & \cmark & 66.49 & 79.33 \\
		\xmark & \cmark & \cmark & 66.71 & 79.86 \\
		\cmark & \xmark & \xmark & 64.09 & 77.51 \\
		\bottomrule
	\end{tabular}
	\label{Table:EffikanComponents}
\end{table}

\smallskip\noindent\textbf{Effect of Bernstein Polynomial Order.}\: Table~\ref{Table:BernsteinOrder} presents the impact of varying the Bernstein polynomial order $R$ in the KAN layer of the AdaptKAN module on segmentation performance, evaluated on the BUSI dataset in terms of IoU and DSC. The global smoothness of feature approximations in AdaptKAN is controlled by the degree of the basis polynomials, which are defined by the Bernstein polynomial order in the KAN layer. Higher orders improve the polynomial's ability to represent complex functions, but because of the additional parameters, there is a chance of overfitting or numerical instability. The results show that a Bernstein polynomial order of three achieves the best performance, striking an optimal trade-off between expressiveness and stability in ultrasound image segmentation due to its ability to capture smooth and continuous lesion boundaries in the BUSI dataset. Reducing the order to two decreases accuracy by 3.26\% in IoU and 2.19\% in DSC, while increasing the order to four results in declines of 2.04\% in IoU and 1.44\% in DSC. Order $R=5$ further reduces performance by 2.74\% in IoU and 1.98\% in DSC. These findings indicate that moderate order Bernstein polynomials provide the best balance between expressive power and over-parameterization.

\begin{table}[!htb]
	\caption{Effect of Bernstein polynomial order inside the KAN layer on AdaKAN's performance evaluated on the BUSI dataset.}
	\smallskip
	\setlength{\tabcolsep}{30pt}
	\centering
	\footnotesize
	\begin{tabular}{@{}ccc@{}}
		\toprule
		Bernstein Order ($R$)  & IoU & DSC \\
		\midrule
		2 &  65.58 & 78.76\\
		3 &  \textbf{68.84} & \textbf{80.95} \\
		4 &  66.80 & 79.51 \\
		5 &  66.10 & 78.97 \\
		\bottomrule
	\end{tabular}
	\label{Table:BernsteinOrder}
\end{table}

\section{Conclusion}
In this paper, we introduced AdaKAN, a novel adaptive framework built upon an encoder-decoder architecture, incorporating an efficient attention that reduces the quadratic computational complexity of the standard attention mechanism to a linear scale, while the adaptive KAN module plays a central role in AdaKAN's adaptability. We conducted extensive experiments across five diverse medical imaging datasets, covering both binary and multi-class segmentation tasks. AdaKAN consistently outperformed state-of-the-art baselines in terms of segmentation accuracy, particularly excelling in boundary delineation and robustness to structural variability. Moreover, qualitative visualizations demonstrated AdaKAN's superior ability to preserve anatomical contours and reduce segmentation artifacts.

\bibliographystyle{ieeetr}
\bibliography{references}
\end{document}